\documentclass{article}
\usepackage[utf8]{inputenc}
\usepackage{authblk}

\makeatletter
\renewcommand\AB@affilsepx{, \protect\Affilfont}
\makeatother
\renewcommand*{\Affilfont}{\small}

\title{Regression on imperfect class labels derived by unsupervised clustering}
\author[1,2]{R.F. Brøndum}
\author[3]{T.Y. Michaelsen}
\author[1,2,4]{M. Bøgsted}
\date{August 2019}

\affil[1]{Department of Haematology, Aalborg University Hospital, Denmark}
\affil[2]{Clinical Cancer Research Center, Aalborg University Hospital, Denmark}
\affil[3]{Center for Microbial Communities, Aalborg University, Denmark}
\affil[4]{Department of Clinical Medicine, Aalborg University, Denmark}

\RequirePackage{natbib}
\RequirePackage{appendix}
\usepackage{graphicx}
\RequirePackage{amsmath}    
\RequirePackage{amssymb}	
\RequirePackage{dsfont} 	
\usepackage{url}

\renewcommand{\vec}{\boldsymbol}

\newcommand{\bbR}{\mathbb{R}}

\newcommand{\bbOne}{\mathds{1}}

\DeclareMathOperator*{\argmax}{arg\,max}
\DeclareMathOperator*{\argmin}{arg\,min}

\let\proglang=\textsf
\let\code=\texttt
\newcommand{\pkg}[1]{{\fontseries{b}\selectfont #1}}

\begin{document}

\maketitle

\begin{abstract}
Outcome regressed on class labels identified by unsupervised clustering is custom in many applications. However, it is common to ignore the misclassification of class labels caused by the learning algorithm, which potentially leads to serious bias of the estimated effect parameters. Due to its generality we suggest to redress the situation by use of the simulation and extrapolation method. Performance is illustrated by simulated data from Gaussian mixture models. Finally, we apply our method to a study which regressed overall survival on class labels derived from unsupervised clustering of gene expression data from bone marrow samples of multiple myeloma patients.
\end{abstract}

\section{Introduction}\label{sec:Introduction}
In biomarker studies it is popular to perform an unsupervised clustering of high-dimensional variables like genome wide screens of SNPs, gene expressions, and protein data and regress for example treatment response, patient recorded outcome measures, time to disease progression, or overall survival on these potentially mislabelled clusters. It is well-known from the statistical literature that errors in continuous and categorical covariates can lead to loss of important information about effects on outcome \citep{Carroll2006}. However, to our surprise this is often ignored when regressing outcome on classes identified by unsupervised learning, which might lead to important clinical effect measures being overlooked \citep{Alizadeh2000d, Veer2002, Guinney2015, Zhan2006, Broyl2010}.       
We suggest to cast the problem as a covariate misclassification problem. This leaves us with a concourse of possible modelling and analysis options, see for example the book by \cite{Carroll2006} or the recent review by \cite{Brakenhoff2018}. A general approach, with good statistical properties, is to maximize the likelihood of a latent variable model joining the regression and classification models \citep{Nevo2016, Skrondal2004}. 

First, this process does not mimic the workflow of the biologists, for whom the basic question is to identify important biological processes and next to relate the clusters to clinical consequences. 
Secondly, upon parameter estimation cluster membership needs post hoc to be estimated by e.g.\ the maximum a posteriori probability, whereby direct connection to the regression parameter is lost. 
Thirdly, this approach requires a statistical model of the clustering process, leaving out the possibility to combine popular unsupervised clustering algorithms, such as hierarchical clustering, with popular parametric regression models like generalized linear models and Cox's proportional hazards model. 

Due to its generality, we chose to study the two-stage modelling process. A number of ad-hoc methods have been developed in various specific settings which could be adapted to this situation. Examples include matrix methods \citep{Morrissey1999}, regression calibration \citep{Rosner2006}, pooled estimation \citep{Spiegelman2001}, multiple imputation \citep{Cole2006}, corrected score estimation \citep{Nakamura1990}, and simulation and extrapolation (simex) \citep{Cook1994, Carroll1996, Carroll1999}. Among these methods, simex has become a useful tool for correcting effect estimates in the presence of additive measurement error. The simex idea has been extended to the misclassification simulation and extrapolation (mcsimex) method for correcting effect estimates in the presence of errors in categorical covariates \citep{Kuchenhoff2006}. In this paper, we chose to focus on mcsimex because of its generality, simplicity, and direct applicability.

The article is organized as follows. In Section \ref{sec:Statistical}, we detail the underlying hierarchical model (Section \ref{subsec:Model}), describe how unsupervised learning and statistical inference are performed (Section \ref{subsec:Inference}), and outline the mcsimex method (Section \ref{subsec:Algorithm}). Simulation studies in Section \ref{sec:Numerical} show the performance of our proposal. An application to a study on unsupervised clustering of gene expression data from bone marrow samples of multiple myeloma patients is given in Section \ref{sec:Example}. The results of the paper is discussed in Section \ref{sec:Discussion} followed by computational details of the the mcsimex method in Section \ref{sec:Computational}. 

\section{Statistical setting}\label{sec:Statistical}

\subsection{The hierarchical model}\label{subsec:Model}
The $m$-component Gaussian mixture model (GMM) of $\vec{Z} = (Z_1, ..., Z_p)^\top$ has the following distribution
\begin{align}
\text{GMM:}\qquad
\begin{cases}
\begin{aligned} \label{GMM}
& H \sim \text{Categorical}(\alpha_1, ..., \alpha_m) \\
& \vec{Z} |H = h \sim \mathcal{N}_p(\vec{\mu}_h, \vec{\Sigma}_h)
\end{aligned}
\end{cases}
\end{align}
where $H \in \{1, 2, ..., m\}$ corresponds to the class and $\alpha_1, ..., \alpha_m, (\geq 0)$ are the mixture proportions satisfying $\sum_{h = 1}^m \alpha_h = 1$. Thus, the GMM is parameterized by
\begin{align*}
\vec{\theta} = (\alpha_1, ..., \alpha_m, \vec{\mu}_1, ..., \vec{\mu}_m, \vec{\Sigma}_1, ..., \vec{\Sigma}_m).
\end{align*}
We denote the possibly misspecified variable $H^*$ for the corresponding correctly specified $H$, i.e.
\begin{align}
  H^* \;|\; H = j \sim \text{Categorical}(\pi_{1j}, ..., \pi_{mj}).    
\end{align}
Thus, the misclassification error is characterized by the $m\times m$ misclassification matrix $\Pi = [\pi_{ij}]_{m\times m}$, which is defined by its components
\begin{align}
  \pi_{ij} = P(H^*=i \;|\; H = j),\; i,j=1,...,m.   
\end{align}
The outcome $Y$ is here modelled as a generalized linear model
\begin{align}
   g(E (Y \;|\; H=h)) = \vec{x}_h^\top\vec{\beta},   
\end{align}
where $\vec{\beta} = (\beta_1,...,\beta_m)^\top$, $\vec{x}_h = \vec{e}_1 + \vec{e}_h \bbOne[h \not = 1]$ is encoded by treatment contrasts, $g$ is the link function, and $Y$ is assumed to be generated from an exponential family distribution \citep{McCullagh1989}. 

\subsection{Statistical inference}\label{subsec:Inference}
Assume we have $n$ i.i.d.\ realizations $\{(y_i,\vec{z}_i), i=1,...,n\}$ of $(Y,\vec{Z})$ and assume we want an estimate of the effect sizes $\vec{\beta}$ of each of the unobserved $m$ components. First, we estimate the parameters of the Gaussian mixture model by maximizing the log-likelihood function to obtain
\begin{align*}
\hat{\vec{\theta}} = (\hat{\alpha_1}, ..., \hat{\alpha_m}, \hat{\vec{\mu}}_1, ..., \hat{\vec{\mu}}_m, \widehat{\vec{\Sigma}}_1, ..., \widehat{\vec{\Sigma}}_m).
\end{align*}
Now, we can estimate the class relationship for any covariate $\vec{z}\in \bbR^p$ by the following hard clustering rule
\begin{align}
\hat{h}(\vec{z}) = \argmax_{h}\hat{p}_h(\vec{z}),
\end{align}
where $\hat{p}_h$ is the density of the $\mathcal{N}_d(\hat{\vec{\mu}}_h, \widehat{\vec{\Sigma}}_h)$-distribution and for the observed covariate we write 
\begin{align}
\hat{h}_i = \hat{h}(\vec{z}_i)
\end{align}
for short.
The misclassification matrix $\Pi$ can be consistently estimated in the following way
\begin{align}
\widehat{\Pi} = \left\{\int \bbOne[\hat{h}(\vec{z}) = i] \hat{p}_j(\vec{z}) d\vec{z}\right\}_{i,j=1}^m.
\end{align}
The next step is to optimize the log-likelihood of the generalized linear model based on the assumed i.i.d.\ data $y_1 | \hat{h}_1,...,y_n | \hat{h}_n$, to obtain the naive estimate $\hat{\vec{\beta}}(\{(y_i,\hat{h}_i), i=1,...,n\})$ \citep{McCullagh1989}. The procedure outlined above will in the following be referred to as the \textit{naive method}.

It is important to notice that maximum likelihood estimation under the naive method is estimation of a misspecified model. Convergence of estimates under misspecified models is ensured by the results of \cite{White1982}. In general we will denote the limit of a maximum likelihood estimate  by $\vec{\beta}(\Pi)$ for a model misspecified by the misclassification matrix $\Pi$.

\subsection{The mcsimex method}\label{subsec:Algorithm}
It is well-known that estimating $\vec{\beta}$ by the naive method leads to a biased estimate \citep{Kuchenhoff2006}. In order to formulate the \textit{mcsimex method} to redress this situation, we define the function $\mathcal{G}: [-1,\infty) \rightarrow \bbR^m$ by
\begin{align}
   \mathcal{G}(\lambda) = \vec{\beta}(\Pi^{(1+\lambda)}), \label{eq:Gfunction} 
\end{align}
where $\Pi^{\lambda}$ can be expressed as $\Pi^{\lambda} = E\Lambda^{\lambda} E^{-1}$ via the spectral decomposition, with $\Lambda$ being the diagonal matrix of eigenvalues and $E$ the corresponding matrix of eigenvectors. For the function (\ref{eq:Gfunction}) to be well-defined, we need to ensure the existence of $\Pi^{\lambda}$ and that it is a misclassification matrix for $\lambda\geq 0$. Criteria for existence are given in \cite{Kuchenhoff2006}.  

We notice that $\mathcal{G}$ parameterizes the amount of misclassification, where $\mathcal{G}(-1)=\vec{\beta}(I_{m\times m})$ corresponds to no misclassfication, $\mathcal{G}(0)=\vec{\beta}(\Pi) $ corresponds to the present misclassification, and $\mathcal{G}(\lambda)$ for $\lambda \geq 0$, corresponds to increasing misclassification. The fundamental idea behind mcsimex is to simulate $\mathcal{G}(\lambda)$ for increasing $\lambda \geq 0$ and then extrapolate back to $\lambda = -1$.  

In a few situations explicit forms of $\mathcal{G}$ as a function of $\lambda$ can be calculated, but they tend to be unstable to estimate, wherefore mcsimex relies on finite-dimensional parametric approximations $\mathcal{G}(\lambda, \gamma)$, $\gamma \in \bbR^k$, of $\mathcal{G}(\lambda)$. One example from the  R-package \pkg{simex} is the quadratic approximation $\mathcal{G}(\lambda, (\gamma_0,\gamma_1,\gamma_2))=\gamma_0+\gamma_1*\lambda+\gamma_2*\lambda^2$ \citep{Lederer2017}. It is custom in the mcsimex litterature to assume one either has the misclassification matrix at hand or it can be estimated from training data. However, in our case the misclassification matrix will be (consistently) estimated from data in the following way
\begin{align}
\widehat{\Pi} = \{\hat{\pi}_{ij}\} = \left\{\int \bbOne[\hat{h}(\vec{z}) = i] \hat{p}_j(\vec{z}) d\vec{z}\right\}.
\end{align}
The mcsimex method (or algorithm) can now be formulated as:
\\
\\
\noindent\textbf{Algorithm \citep{Kuchenhoff2006}}
\begin{enumerate}
    \item For a fixed grid of values $1 < \lambda_1 <\cdots <\lambda_k$, simulate i.i.d.\ random variables (condition upon $\{(y_i,\vec{z}_i), i=1,...,n\})$ 
    \begin{align*}
        H^*_{b,i}(\lambda_k) \sim \text{Categorical}\left(\widehat{\Pi}^{\lambda_k}_{\bullet \hat{h}_i}\right),
    \end{align*}
    where $b=1,...,B$, $i=1,...,n$, $k=1,...,m$, and $\widehat{\Pi}^{\lambda_k}_{\bullet \hat{h}_i}$ is the $\hat{h}_i$'th column of $\widehat{\Pi}^{\lambda_k}$.
    \item Set
    \begin{align*}
        \hat{\vec{\beta}}({\lambda_k}) = B^{-1}\sum_{b=1}^B  \hat{\vec{\beta}}(\{(y_i,h^*_{b,i}(\lambda_k)), i=1,...,n\}).
    \end{align*}
    \item Estimate $\gamma$ by the least squares method
    \begin{align*}
        \hat{\gamma} = \argmin_{\gamma \in \Gamma} \sum_{i=0}^{k} \left(\hat{\vec{\beta}}({\lambda_k}) - \mathcal{G}(\lambda_k, \gamma)\right)^2,
    \end{align*}
    where $\lambda_0 = 1$ and $\hat{\vec{\beta}}({\lambda_0}) = \hat{\vec{\beta}}(\{(y_i,\hat{h}_i), i=1,...,n\})$ is the naive estimator.
    \item The mcsimex estimate is then given by the extrapolation to $\lambda = -1$
    \begin{align*}
        \hat{\vec{\beta}}_{S} = \mathcal{G}(-1,\hat{\gamma}).
    \end{align*}
\end{enumerate}

\section{Simulation studies}\label{sec:Numerical}
\subsection{Logistic regression}
In this section, we investigate empirically how different sample sizes, imbalances between classes, and clustering algorithms affect the effect estimates. For the simulations, we generate $n = 200$, $n=500$, or $n=1000$ independent samples from a 2-class Gaussian mixture model, where the prior probabilities of class 1 is $\pi_1 = 5/10$ or $\pi_1 = 2/10$ to generate balanced or imbalanced classes, respectively, and set $\pi_2 = 1 - \pi_1$. Class 1 and 2 observations have bivariate normal distributions with means $\mu_1=(-1, 0)$ and $\mu_2=(1, 0)$ respectively, and a common identity covariance matrix. The outcome is modelled by a logistic regression with linear predictor $\vec{x}_h^\top\vec{\beta}$ where the intercept and class effect are given by $(\beta_1, \beta_2) = (-1,2)$.

Unsupervised clustering was performed using either Gaussian mixture models as implemented in the R-package {\bf mcclust} \citep{scrucca2016}, or $k$-means from the base implementation in R. We estimated the parameter $\vec{\beta}$ using either the true class labels, class labels inferred from unsupervised clustering, and the mcsimex method. The misclassification matrix for the mcsimex approach was inferred by drawing $100,000$ bivariate normal samples using the means and variance-covariance matrix from each of the inferred clusters and counting how often they were misclassified by the fitted clustering model.

Each scenario was repeated 1000 times and the results are summarized by bias, mean-square error, and coverage of the estimated confidence intervals. Results from the balanced scenario $\pi_1 = 5/10$ are shown in Table \ref{simResults:balanced}. We see that using the true class labels there is a small bias in both of the estimated parameters and the coverage is close to $95\%$. When using the estimated class labels inferred from either the GMM or K-means clustering without taking misclassification into account, i.e. the naïve model, both parameters are biased and coverage is far from the assumed $95\%$. The bias is not alleviated by increasing the number of samples, the coverage is however smaller, due to a smaller standard error. When taking misclassification into account, using the simex approach, both bias and coverage of the parameters is improved, and the improvement is similar across sample sizes. Results from the imbalanced scenario are shown in Table \ref{simResults:unbalanced}. These also show smaller bias and better coverage for the simex model. Results are, however, better when using GMM than $k$-means. The $k$-means algorithm tends to estimate clusters of uniform size \citep{HuiXiong2009}, leading to poor performance with imbalanced clusters, and since we used the fitted $k$-means clusters to infer the misclassification matrix this also becomes misspecified and the simex approach cannot fully correct for the added noise. Some work has been done to alleviate this bias in the $k$-means algorithm, e.g. using multicenters \citep{Liang2012} or undersampling \citep{Kumar2014}. We did not pursue this further in the present paper, but just notice that one might instead use the GMM to infer clusters in the imbalanced case. 

\begin{table}[!tbp]
	\caption{Results from applying simex to simulated data with balanced classes. A logistic regression model was fitted with the true or inferred class labels, from either gaussian mixture models (GMM) or $k$-means with and without simex correction. Simulations were done with $200, 500$, or $1000$ samples.
		\label{simResults:balanced}} 
	\begin{center}
		\begin{tabular}{llcllcll}
			\hline\hline
			\multicolumn{1}{l}{\bfseries }&\multicolumn{1}{c}{\bfseries }&\multicolumn{1}{c}{\bfseries }&\multicolumn{2}{c}{\bfseries GMM}&\multicolumn{1}{c}{\bfseries }&\multicolumn{2}{c}{\bfseries Kmeans}\tabularnewline
			\cline{4-5} \cline{7-8}
			\multicolumn{1}{l}{}&\multicolumn{1}{c}{True}&\multicolumn{1}{c}{}&\multicolumn{1}{c}{Naive}&\multicolumn{1}{c}{Simex}&\multicolumn{1}{c}{}&\multicolumn{1}{c}{True}&\multicolumn{1}{c}{Naive}\tabularnewline
			\hline
			{\bfseries 200}&&&&&&&\tabularnewline
			~~bias $\beta_1$&-0.01&&0.34&0.07&&0.34&0.07\tabularnewline
			~~bias $\beta_2$&0.03&&-0.69&-0.17&&-0.67&-0.12\tabularnewline
			~~coverage $\beta_1$&0.94&&0.55&0.86&&0.6&0.92\tabularnewline
			~~coverage $\beta_2$&0.94&&0.39&0.9&&0.39&0.93\tabularnewline
			\hline
			{\bfseries 500}&&&&&&&\tabularnewline
			~~bias $\beta_1$&0&&0.36&0.08&&0.34&0.06\tabularnewline
			~~bias $\beta_2$&0&&-0.71&-0.15&&-0.69&-0.14\tabularnewline
			~~coverage $\beta_1$&0.96&&0.28&0.88&&0.28&0.92\tabularnewline
			~~coverage $\beta_2$&0.96&&0.06&0.91&&0.06&0.9\tabularnewline
			\hline
			{\bfseries 1000}&&&&&&&\tabularnewline
			~~bias $\beta_1$&0&&0.35&0.06&&0.35&0.08\tabularnewline
			~~bias $\beta_2$&0.01&&-0.69&-0.12&&-0.69&-0.15\tabularnewline
			~~coverage $\beta_1$&0.95&&0.08&0.88&&0.05&0.87\tabularnewline
			~~coverage $\beta_2$&0.96&&0&0.9&&0&0.88\tabularnewline
			\hline
	\end{tabular}\end{center}
\end{table}

\begin{table}[!tbp]
	\caption{Results from applying simex to simulated data with imbalanced classes. A logistic regression model was fitted with the true or inferred class labels, from either Gaussian mixture models (GMM), or $k$-means with and without simex correction. Simulations were done with $200, 500$ or $1000$ samples.\label{simResults:unbalanced}} 
	\begin{center}
		\begin{tabular}{llcllcll}
			\hline\hline
			\multicolumn{1}{l}{\bfseries }&\multicolumn{1}{c}{\bfseries }&\multicolumn{1}{c}{\bfseries }&\multicolumn{2}{c}{\bfseries GMM}&\multicolumn{1}{c}{\bfseries }&\multicolumn{2}{c}{\bfseries Kmeans}\tabularnewline
			\cline{4-5} \cline{7-8}
			\multicolumn{1}{l}{}&\multicolumn{1}{c}{True}&\multicolumn{1}{c}{}&\multicolumn{1}{c}{Naive}&\multicolumn{1}{c}{Simex}&\multicolumn{1}{c}{}&\multicolumn{1}{c}{True}&\multicolumn{1}{c}{Naive}\tabularnewline
			\hline
			{\bfseries 200}&&&&&&&\tabularnewline
			~~bias $\beta_1$&-0.06&&0.52&0.14&&1.09&0.83\tabularnewline
			~~bias $\beta_2$&0.06&&-0.7&-0.17&&-1.16&-0.74\tabularnewline
			~~coverage $\beta_1$&0.96&&0.55&0.81&&0.01&0.25\tabularnewline
			~~coverage $\beta_2$&0.95&&0.49&0.89&&0.05&0.61\tabularnewline
			\hline
			{\bfseries 500}&&&&&&&\tabularnewline
			~~bias $\beta_1$&0&&0.57&0.14&&1.09&0.82\tabularnewline
			~~bias $\beta_2$&0.01&&-0.75&-0.16&&-1.15&-0.73\tabularnewline
			~~coverage $\beta_1$&0.95&&0.32&0.82&&0&0.04\tabularnewline
			~~coverage $\beta_2$&0.95&&0.2&0.89&&0&0.32\tabularnewline
			\hline
			{\bfseries 1000}&&&&&&&\tabularnewline
			~~bias $\beta_1$&-0.01&&0.53&0.08&&1.08&0.81\tabularnewline
			~~bias $\beta_2$&0.01&&-0.71&-0.1&&-1.14&-0.72\tabularnewline
			~~coverage $\beta_1$&0.94&&0.16&0.86&&0&0\tabularnewline
			~~coverage $\beta_2$&0.95&&0.04&0.89&&0&0.08\tabularnewline
			\hline
	\end{tabular}\end{center}
\end{table}

\subsection{Cox Proportional Hazards}
In survival analysis the Cox proportional hazards model is often used as the model of choice for inferring the impact of covariates on the rate of events. The R package {\bf simex}, used for this paper, did not previously support this model, which poses a problem for using mcsimex in survival analysis. This can be circumvented by using the Poisson approximation as done in \cite{Bang2013}, but we chose instead to augment the source code of the \pkg{simex} package to include the coxph model class from the \pkg{survival} package \citep{survival-package, survival-book}. 

To test the performance of our implementation we performed a second round of simulations. Here we simulated $200, 500$, or $1000$ samples, where class labels, $(0, 1)$, were drawn from a binomal distribution with parameter $\pi = 0.5$ and added misclassification noise at a rate of $0.1, 0.2$, or $0.3$ (off diagonal values in the misclassification matrix). Survival times were drawn from an exponential distribution with parameter $\lambda = \text{class} + 1$ and censoring times were drawn from and exponential distribution with parameter $\lambda = 0.5$. The results from 1000 simulations of each scenario are shown in Table \ref{simResults:cox}. Results confirm that the simex method also reduces biases in the estimated parameters from the Cox proportional hazards model. However, the improvement in bias as well as the coverage is better for lower rates of misclassification. The coverage indicates that standard errors for the estimates are too small, which is likely caused by estimating them with the jackknife variance estimator in the \pkg{simex} package. This has previously been shown to underestimate the variance, so using an asymptotic variance estimate is preferred \citep{Kuchenhoff2007}. An asymptotic variance estimate for the Cox proportional hazards model, incorporating misclassificaion, has to the best of our knowledge not yet been derived, but the problem can be alleviated by using a bootstrap approach to estimate the variance, which also adds the possibility to include additional variance resulting from estimation of the misclassification matrix \citep{Kuchenhoff2007}. However, the bootstrap approach drastically increases computational cost as the simex model has to be fitted for each bootstrap sample.

\begin{table}[!tbp]
	\caption{Results from cox estimation of 1000 simulations using a misclassification probability (mp) of 0.1, 0.2 or 0.3\label{simResults:cox}} 
	\begin{center}
		\begin{tabular}{lrcrrcrrcrr}
			\hline\hline
			\multicolumn{1}{l}{\bfseries }&\multicolumn{1}{c}{\bfseries }&\multicolumn{1}{c}{\bfseries }&\multicolumn{2}{c}{\bfseries mp = 0.1}&\multicolumn{1}{c}{\bfseries }&\multicolumn{2}{c}{\bfseries mp = 0.2}&\multicolumn{1}{c}{\bfseries }&\multicolumn{2}{c}{\bfseries mp = 0.3}\tabularnewline
			\cline{4-5} \cline{7-8} \cline{10-11}
			\multicolumn{1}{l}{}&\multicolumn{1}{c}{True}&\multicolumn{1}{c}{}&\multicolumn{1}{c}{Naive}&\multicolumn{1}{c}{Simex}&\multicolumn{1}{c}{}&\multicolumn{1}{c}{Naive}&\multicolumn{1}{c}{Simex}&\multicolumn{1}{c}{}&\multicolumn{1}{c}{Naive}&\multicolumn{1}{c}{Simex}\tabularnewline
			\hline
			{\bfseries 200}&&&&&&&&&&\tabularnewline
			~~bias&$0.06$&&$-0.23$&$ 0.07$&&$-0.48$&$-0.03$&&$-0.68$&$-0.22$\tabularnewline
			~~coverage&$0.95$&&$ 0.87$&$ 0.94$&&$ 0.60$&$ 0.93$&&$ 0.29$&$ 0.83$\tabularnewline
			\hline
			{\bfseries 500}&&&&&&&&&&\tabularnewline
			~~bias&$0.01$&&$-0.27$&$ 0.00$&&$-0.50$&$-0.10$&&$-0.70$&$-0.30$\tabularnewline
			~~coverage&$0.94$&&$ 0.71$&$ 0.94$&&$ 0.22$&$ 0.88$&&$ 0.02$&$ 0.74$\tabularnewline
			\hline
			{\bfseries 1000}&&&&&&&&&&\tabularnewline
			~~bias&$0.01$&&$-0.28$&$-0.01$&&$-0.51$&$-0.11$&&$-0.69$&$-0.31$\tabularnewline
			~~coverage&$0.95$&&$ 0.48$&$ 0.93$&&$ 0.04$&$ 0.86$&&$ 0.00$&$ 0.64$\tabularnewline
			\hline
	\end{tabular}\end{center}
\end{table}

\section{Cancer sub-classification}\label{sec:Example}
Since the invention of high dimensional gene expression profiling, just before this millennium, a popular task has been to perform unsupervised cluster analysis on such data to identify new subgroups and correlate these subgroups to biological information, clinical data, and outcome. In this paper, we consider an example from sub-classification of multiple myleoma (MM). MM is a malignancy of end stage B cells that expand in the bone marrow, resulting in anemia, bone destruction, and renal failure. The data set contains 414 gene expressions sets from multiple myeloma patient's bone marrow \citep{Zhan2006}. The gene expressions were profiled on Affymetrix HGU133 Plus 2 arrays and exported to .CEL-files by the Affymetrix Genomics Console. To replicate the analysis of \cite{Zhan2006} we downloaded raw .CEL files from the GEO repository GS24080 and matched these by patient IDs to cases included in \cite{Zhan2006}, and were able to match 407 out of the 414 cases. This was done since raw data were not available in the repository indicated in the original paper, so we resorted to a later study from the same group. Data were MAS5 normalized and filtered according to instructions from \cite{Zhan2006} resulting in a dataset with gene expressions for 2,169 genes. This is a higher number of genes than the 1,559 reported in the original paper, and is possibly caused by the different number of samples and/or slight differences in the MAS5 normalization procedure as implemented in the R package {\bf affy} \citep{Gautier2004} and the Affymetrix Microarray Suite GCOS 1.1. 

\begin{figure}
    \centering
    \includegraphics[width=120mm]{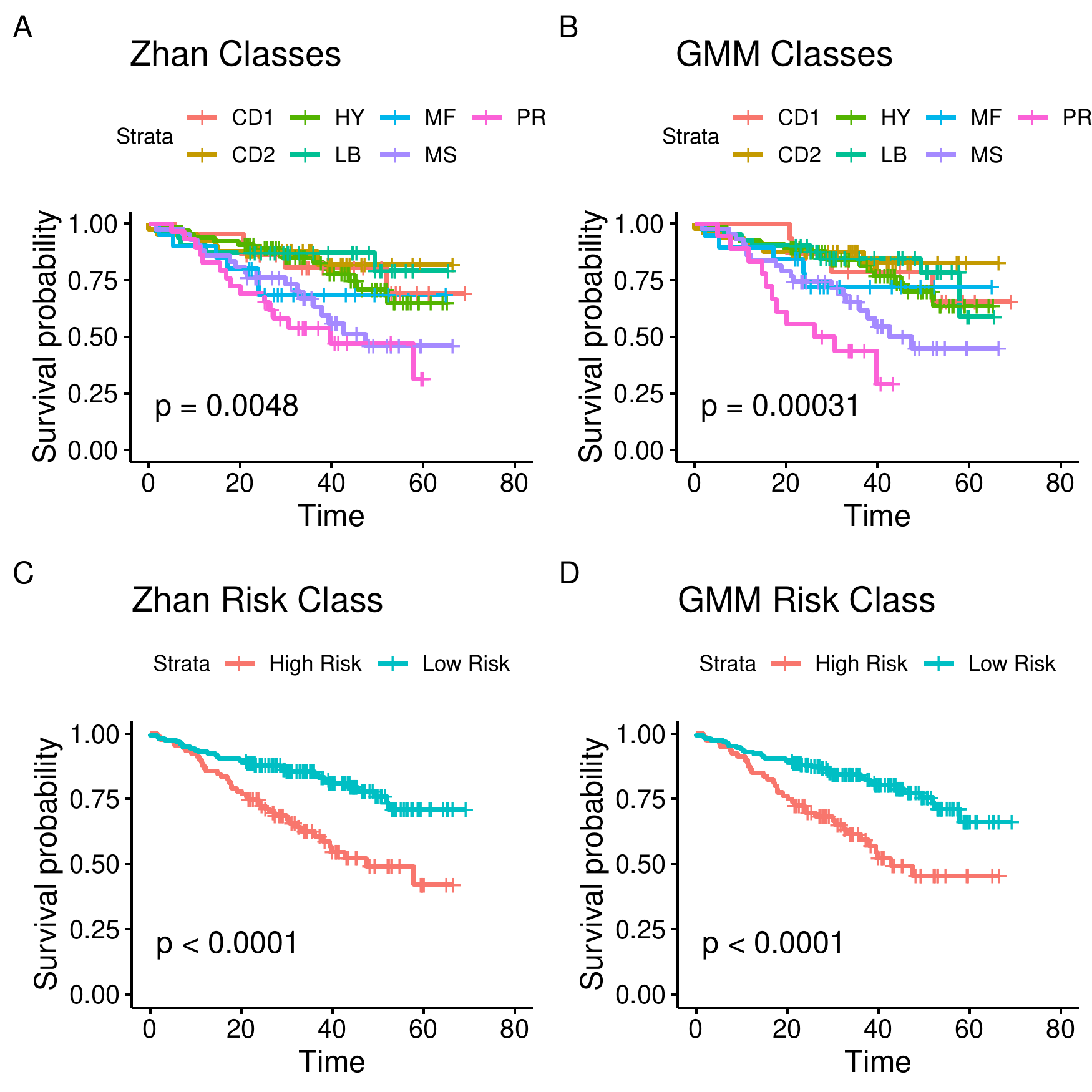}
    \caption{Survival curves for the identified classes, and risk groups in the original study and refitted with GMM respectively.}
    \label{fig:mm.surv}
\end{figure}

After filtering, they performed hierarchical clustering on the remaining genes and chose to cut the tree, so seven clusters were formed. In order to mimic their analysis flow we identified seven sub-groups by estimating a 7-component GMM. The classes were labelled according to the most similar class from \cite{Zhan2006} as shown in the confusion matrix in Table \ref{confmm:train}. The classes estimated from the GMM had an accuracy of 0.9 compared to the original classes. Survival curves are shown for the original classes and GMM classes in respectively panels A and B of Figure \ref{fig:mm.surv}. These data confirm, that a Gaussian mixture model reasonably well approximates the hierarchical clustering of \cite{Zhan2006}.

\begin{table}[!tbp]
	\caption{Confusion matrix for training set of GSE4581, accuracy =  0.9. Rows show original classes from \cite{Zhan2006} and columns show clusters determined with the Gaussian mixture model.\label{confmm:train}} 
	\begin{center}
		\begin{tabular}{lrrrrrrr}
			\hline\hline
			\multicolumn{1}{l}{}&\multicolumn{1}{c}{1}&\multicolumn{1}{c}{2}&\multicolumn{1}{c}{3}&\multicolumn{1}{c}{4}&\multicolumn{1}{c}{5}&\multicolumn{1}{c}{6}&\multicolumn{1}{c}{7}\tabularnewline
			\hline
			CD1&$15$&$ 6$&$ 0$&$ 0$&$ 0$&$ 1$&$ 0$\tabularnewline
			CD2&$ 0$&$40$&$ 0$&$ 0$&$ 0$&$ 0$&$ 1$\tabularnewline
			HY&$ 0$&$ 2$&$63$&$ 0$&$ 0$&$ 0$&$ 0$\tabularnewline
			MS&$ 0$&$ 0$&$ 0$&$42$&$ 0$&$ 0$&$ 0$\tabularnewline
			MF&$ 0$&$ 0$&$ 0$&$ 0$&$19$&$ 1$&$ 0$\tabularnewline
			PR&$ 0$&$ 0$&$ 1$&$ 1$&$ 0$&$16$&$11$\tabularnewline
			LB&$ 0$&$ 0$&$ 2$&$ 0$&$ 0$&$ 0$&$29$\tabularnewline
			\hline
	\end{tabular}\end{center}
\end{table}

The 7 classes in \cite{Zhan2006} were split into a low risk group consisting of the CD1, CD2, HY, and LB classes, and a high risk group with the MF, MS, and PR classes with survival differences as shown in panel C of Figure \ref{fig:mm.surv} for the original classes and panel D for the GMM classes. A Cox proportional hazards model was employed to estimate the hazard ratio of high vs low risk group, but this analysis did not take any possible misclassification of the inferred classes from the unsupervised clustering into account. To investigate the impact of correction for misclassification we applied the mcsimex method to the data at hand. As shown in the simulation results the variance estimates of the built-in jackknife estimate are underestimated so we chose to do the analysis using bootstraps as well. We performed 1000 bootstraps iterations where at each step a sample of size $n=407$ was drawn with replacement from the available data. A GMM was fitted to the sample and the out-of-bag samples was used to infer the misclassification matrix by comparing the predicted class from the in-bag GMM model to the class obtained from the full data. A Cox proportional hazards model was then fitted to the in-bag sample and the mcsimex model was applied using the estimated misclassification matrix. By estimating the misclassification matrix at each iteration of the bootstrap procedure we factor in the added variance from its estimation. Results from this procedure,  compared to the naive estimate along with results from using mcsimex with an average misclassification matrix from 1000 bootstraps, but only fitting the mcsimex model once, are shown in Table \ref{tab.cox.mm}. For the average misclassification matrix we observed a misclassification probability of $0.07$ for the low risk group, and $0.13$ for the high risk group. For the naïve models we see similar results for the original classes from \cite{Zhan2006} and the refitted classes from the GMM, and both approaches for correction for misclassification gives a higher point estimate for the hazard ratio of high vs low risk, confidence intervals are, however, wide and overlapping.

\begin{table}[!tbp]
	\caption{Hazard ratio of high vs low risk groups in the training set from \cite{Zhan2006} with the naïve and simex corrected models using the GMM classes.\label{tab.cox.mm}} 
	\begin{center}
		\begin{tabular}{lrrrr}
			\hline\hline
			\multicolumn{1}{l}{}&\multicolumn{1}{c}{HR}&\multicolumn{1}{c}{Lower 95}&\multicolumn{1}{c}{Upper 95}&\multicolumn{1}{c}{P-value}\tabularnewline
			\hline
			Zhan - Naïve&$2.53$&$1.58$&$4.06$&$1.19e-04$\tabularnewline
			GMM - Naïve&$2.55$&$1.59$&$4.09$&$9.57e-05$\tabularnewline
			GMM - Simex (Average MC)&$3.15$&$1.76$&$5.66$&$1.53e-04$\tabularnewline
			GMM - Simex (Full bootstrap)&$3.12$&$1.72$&$5.66$&$2.28e-04$\tabularnewline
			\hline
	\end{tabular}\end{center}
\end{table}

\section{Discussion}\label{sec:Discussion}
In this paper, we documented a bias on effect estimates when regressing on misclassfied labels arising from unsupervised learning. We also suggested a workflow for adjusting the effect estimates based on the mcsimex method. We had to extend existing software to appropriately handle regression based on time to event outcome. The effectiveness of the workflow was documented on simulated data and we also shed new light on bias and variance of effect estimates in an existing cancer sub-classification study. 

The strength of the suggested workflow is essentially its general applicability and seemingly robustness. However, these advantages come at the cost of computational intensiveness of the Monte Carlo approach in mcsimex.  

As mentioned in Section \ref{sec:Introduction} there exists a number of alternative methods to handle misclassifed labels in regression models. Latent variable models seem most interesting as they are built on parametric models and maximum likelihood estimation. \cite{Nevo2016} is the only paper we have come across dealing with the misclassification problem arising from unsupervised learning. In this paper, though, a slightly different set-up is studied, as they formulate a latent variable model for class risk given measured class features and additional covariates. 

In the light of our results, we encourage researchers to adjust for bias when regressing on potentially misclassified labels. We also encourage biomarker researchers to re-visit previous studies, especially those, which led to negative results when regressing upon misclassified labels. 

\section{Computational details}\label{sec:Computational}
All simulations and analyses were carried out by the statistical programming language \proglang{R}, using the \code{mcsimex} function from the \pkg{simex} package as the main vehicle \citep{Lederer2017}. This function contains functionality for mcsimex correction of regressions from the \code{lm}, \code{glm}, \code{gam}, \code{nls}, \code{polr}, \code{lme}, and \code{nlme} functions. For the current study we extended the package to accomodate Cox proportional hazards regression via the \code{coxph} function of the \pkg{survival} package \citep{survival-book, survival-package}. This extension was made by forking the source code of the \pkg{simex} package from \url{https://github.com/cran/simex}. The adapted code has been included in the \pkg{simex} package and is available at \url{https://cran.r-project.org/package=simex}

We utilized a number of other R and Bioconductor packages, notably the \pkg{mclust} package for clustering \citep{scrucca2016}. For a complete list of packages see the Rmarkdown document available at \url{https://github.com/HaemAalborg/misClass}, which details all steps in the analyses carried out in this paper.

\section{Acknowledgements}
Part of the work was done while Martin Bøgsted visited University of Cambridge. He wants to thank Silvia Richardson, Rajen Shah, and Richard Samworth for their comments on the work and hospitality during the visit and the North Denmark Region's Health Scientific Research Fund and the Lundbeck Foudation for financial support.

\bibliographystyle{apalike}
\bibliography{misclass}

\end{document}